\documentclass{article}

\usepackage[nonatbib,preprint]{neurips_2024}

\usepackage{array}
\newcolumntype{C}[1]{>{\centering\arraybackslash}p{#1}}

\usepackage{graphicx}
\usepackage{booktabs}
\usepackage{multirow}
\usepackage{array}
\usepackage{amssymb}
\usepackage{amsmath}

\DeclareUnicodeCharacter{0301}{\'{e}}

\usepackage{makecell}

\usepackage{floatrow}
\usepackage{tabularx}

\usepackage[utf8]{inputenc} 
\usepackage[T1]{fontenc}    
\usepackage{hyperref}       
\usepackage{url}            
\usepackage{booktabs}       
\usepackage{amsfonts}       
\usepackage{nicefrac}       
\usepackage{microtype}      
\usepackage{xcolor}         

\usepackage{hyperref}
\usepackage{graphicx}
\usepackage{rotating}

\usepackage{orcidlink}
\usepackage{multirow}
\usepackage{mathrsfs}
\usepackage{amsmath,amssymb}
\usepackage{makecell}
\usepackage[ruled,noend,linesnumbered]{algorithm2e}

\usepackage{enumitem}

\usepackage{wrapfig}

\usepackage{caption}
\usepackage{subcaption}

\usepackage{floatrow}
\newfloatcommand{capbtabbox}{table}[][\FBwidth]

\usepackage[
backend=biber,
sorting=anyt,
bibstyle=ieee,
citestyle=numeric,
]{biblatex}
\title{Diffusion Model Meets Non-Exemplar Class-Incremental Learning and Beyond}

\author{%
  Jichuan Zhang, Yali Li\footnotemark[1], Xin Liu, Shengjin Wang\\
  Beijing National Research Center for Information Science and Technology (BNRist)\\
Department of Electronic Engineering, Tsinghua University
\\
  \texttt{\{zhangjc22, xinliu20\}@mails.tsinghua.edu.cn}\\
  \texttt{\{liyali13, wgsgj\}@tsinghua.edu.cn}\\ 
}

\begin{document}

\maketitle
\renewcommand{\thefootnote}{\fnsymbol{footnote}} 
\footnotetext[1]{Corresponding author} 

\begin{abstract}
  Non-exemplar class-incremental learning (NECIL) is to resist catastrophic forgetting without saving old class samples.  Prior methodologies generally employ simple rules to generate features for replaying, suffering from large distribution gap between replayed features and real ones. To address the aforementioned issue, we propose a simple, yet effective \textbf{Diff}usion-based \textbf{F}eature \textbf{R}eplay (\textbf{DiffFR}) method for NECIL. First, to alleviate the limited representational capacity caused by fixing the feature extractor, we employ Siamese-based self-supervised learning for initial generalizable features. Second, we devise diffusion models to generate class-representative features highly similar to real features, which provides an effective way for exemplar-free knowledge memorization. Third, we introduce prototype calibration to direct the diffusion model's focus towards learning the distribution shapes of features, rather than the entire distribution. Extensive experiments on public datasets demonstrate significant performance gains of our DiffFR, outperforming the state-of-the-art NECIL methods by 3.0\% in average. The code will be made publicly available soon.
\end{abstract}

\section{Introduction}
\label{sec:intro}

Based on pre-collected datasets such as large-scale images or texts, deep neural networks based models on fixed benchmarks have achieved significant success in various fields. However, in practical world, it is common that data arrives continuously in a streaming format \cite{gomes2017survey}. Consequently, the models are required be updated incrementally to fit the continually emerged data. Fine-tuning the model directly on new data significantly suffers from \textit{catastrophic forgetting}. In contrast, joint training on old and new data is expensive in both computation and storage. Recently, several class-incremental learning (CIL) methods \cite{castro2018end, hou2019learning, rebuffi2017icarl, wu2019large, zhao2020maintaining, yan2021dynamically} rely on the data storage to alleviate the so-called \textit{stability-plasticity dilemma}. Yet due to the privacy concerns \cite{venkatesan2017strategy} and memory limitations \cite{ravaglia2021tinyml}, non-exemplar class-incremental learning (NECIL) without the requirement of available old data has attracted increasing attention in both research and real applications \cite{yu2020semantic, smith2021always, zhu2021class, zhu2021prototype}.

\begin{figure}[t]
  \centering
  \includegraphics[scale=0.18]{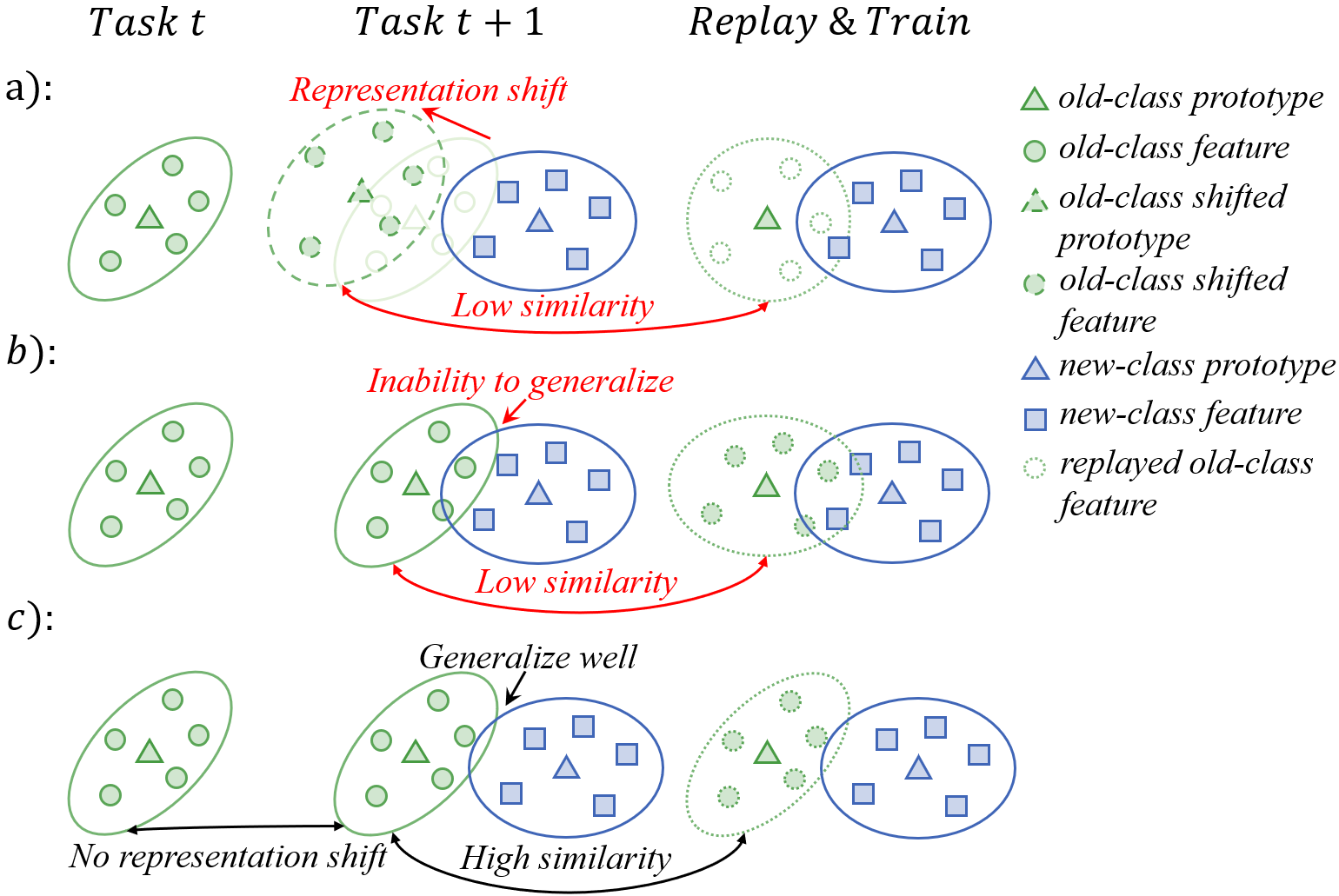}
  \caption{Comparisons of three feature replay ways. a) and b) represent methods exemplified by PASS and FeTrIL, respectively. c) represent our method (DiffFR). Compared to a) and b), we replay features with high similarity to real features, thus improving the performance of NECIL. Besides,our approach does not suffer from representation shift or the inability to generalize.}
  \label{fig:intro}
\end{figure}

Existing NECIL methods incorporate feature replay in various manners. The knowledge distillation-based methods like PASS \cite{zhu2021prototype} regularize the parameter changing between feature extractor on new classes and old prototypes. Since the feature extractor is sequentially updated, the real features corresponding to old classes also undergo changes, which would lead to the representation shift, as shown in Fig.\ref{fig:intro}(a). The transfer learning-based methods like FeTrIL \cite{petit2023fetril} train the feature extractor on the initial data and freeze it afterwards to maintain the stability (and tackle the forgetting issue). Yet the generalization ability on new classes is restricted and highly depends on the initial supervised learning, as shown in Fig.\ref{fig:intro}(b). There are other methods which involves prototype-based simple rules or consistency of statistical values between generated and real features. However, the considerable discrepancy in representation distribution persists, leading to performance degradation in class-incremental learning.

In this paper, we propose \textbf{Diff}usion-based \textbf{F}eature \textbf{R}eplay, abbreviated as \textit{\textbf{DiffFR}}, for non-exemplar class-incremental learning. Inspired by the powerful generative ability of diffusion models \cite{dhariwal2021diffusion}, we propose to generate class-representative replaying features with conditional diffusion models. To balance between the feature representativeness and memory consumption, we specially design a one-dimensional UNet and the attached diffusion model for effective and compact knowledge preserving. 
Besides, self-supervised learning is employed in the initial training to train the feature extractor with high generalization ability, which would help reduce the representation overlap and confusion in feature space. Moreover, we propose prototype calibration mechanism to restrain the diffusion models focusing on feature distribution shape. As the result, our proposed \textit{\textbf{DiffFR}} can resist the representation shift and generate features with high similarity, as shown in Fig.\ref{fig:intro}(c). It also provides an effective way to preserve the knowledge of old classes and promote the generalization on new classes.

Generally, we circumvent the catastrophic forgetting from the perspective of generative models. We adopt the strategy of freezing feature extractor after initial training for stability and incrementally updating the classifier for plasticity. Our proposed \textit{DiffFR} is characterized by three properties. First, we integrate the self-supervised learning with instance similarity constraints to train the feature extractor. By taking the class- and instance-level representation into account, the overlap and confusion between feature distributions can be reduced, thereby promoting the generalization ability to unseen incremental classes. Second, diffusion models are introduced to fit the feature distribution and generate diverse features similar to real ones. Those generated features act as old class prototypes for feature replay, which is effective and efficient for memorization. Notably, our DiffFR differs from existing diffusion based CIL methods \cite{gao2023ddgr, jodelet2023class} that it generates features instead of images, with the core that feature is the compact representations for memorization and replaying. Third, prototype calibration is further proposed to empower the diffusion process by enhancing the learning of feature distribution. The main contributions are summarized as follows:

\begin{itemize}
\item We propose to integrate Siamese-based self-supervised learning for generalizable features. By constraints on class- and instance-level discrimination, the feature overlaps can be reduced for generalization and transferring. 

\item We propose a novel NECIL approach named DiffFR that builds diffusion models for feature replay. By replaying at the feature level rather than the image level, we achieve improvements in both performance and efficiency.

\item We introduce prototype calibration forcing diffusion models to fit the centralized distributions, which can strength the diffusion process for class-aware feature generation.  

\end{itemize}

Extensive experiments on public benchmarks including CIFAR-100, TinyImageNet and ImageNet-Subset demonstrate the superiority of our proposed DiffFR. It achieves significant performance improvements over existing NECIL methods, by average 3.0\%. It also provide the new state-of-the-art for NECIL methods. 

\section{Related Work}
\label{sec:Related_Work}

\textbf{Class-Incremental Learning.} Existing CIL methods can be broadly divided into three categories: rehearsal-based, regularization-based and structure-based methods. 
\textit{Rehearsal-based methods} preserve knowledge of old classes by storing a fixed-size memory of exemplars. Some works \cite{li2017learning,rebuffi2017icarl,wu2019large,smith2021always,hou2019learning, douillard2020podnet} perform knowledge distillation to build the mapping between old and new models, while others \cite{lopez2017gradient, chaudhry2018efficient, wang2021training} try to regularize the model with former data and control the optimization direction. \textit{Regularization-based methods} \cite{kirkpatrick2017overcoming, zenke2017continual, aljundi2018memory, paik2020overcoming} posit varying importance of different parameters within a network. Employing diverse estimation techniques to quantify parameter importance, they penalize updates to crucial parameters to maintain previous knowledge. \textit{Structure-based methods} \cite{yoon2017lifelong, hung2019compacting, li2019learn, yan2021dynamically, wang2022foster, zhou2022model} contend that adapting to new features in a model with limited capacity will lead to the overwriting and forgetting of old features. They dynamically expand the model’s representation ability, such as backbone expansion and prompt expansion, to fit the evolving data stream.

Many CIL approaches rely on the saving exemplars to get better performance, which is limited by memory or privacy issues. In contrast, NECIL does not store any exemplars from past classes, which is more practical. Many recent NECIL methods \cite{zhu2021prototype,zhu2021class,zhu2022self} integrate past class prototypes with distillation to enhance performance, generally favor plasticity. PASS \cite{zhu2021prototype} proposes prototype augmentation to improve the discriminative capability of learned classes across different incremental states. IL2A \cite{zhu2021class} leverages class distribution information to replay features of past classes. SSRE \cite{zhu2022self} introduces a prototype selection mechanism to reduce confusion between new classes and old classes.
However, they suffer from the representation shift caused by the sequential updating of the feature extractor. Previous comparative studies \cite{belouadah2021comprehensive, masana2022class} have revealed that, while distillation-based approaches are theoretically appealing, they exhibit subpar performance in NECIL, particularly for large-scale datasets. Inspired by transfer learning \cite{neyshabur2020being}, some other works \cite{belouadah2018deesil, petit2023fetril} train the feature extractor on the initial data and then keep it fixed to avoid the representation shift. They have shown impressive performance and note that our method falls into this category.

\textbf{Diffusion Model.} Diffusion models draw inspiration from non-equilibrium thermodynamics, and they have shown exceptional performance in various image-related tasks, including unconditional and class-conditional generation \cite{dhariwal2021diffusion}, image-to-image translation \cite{saharia2022image} and text-to-image synthesis \cite{rombach2022high}. DDPM \cite{ho2020denoising} constructs a Markov chain of discrete steps to progressively add random noise for the input and then learn to reverse the diffusion process, thereby enabling the generation of desired data samples from the noise. It exhibits slow sampling speed and DDIM \cite{song2020denoising} can be used to accelerate the denoising process.
Conditional generation is primarily achieved through two approaches: classifier guidance \cite{dhariwal2021diffusion, liu2023more} and classifier-free guidance \cite{ho2022classifier}. The former involves initially training an unconditional diffusion model, followed by a classifier to guide generation process. The latter incorporates conditions directly during training. Here we adopt the simple yet effective classifier-free guidance approach for class-conditional generation.

\begin{figure*}
  \centering
  \includegraphics[scale=0.22]{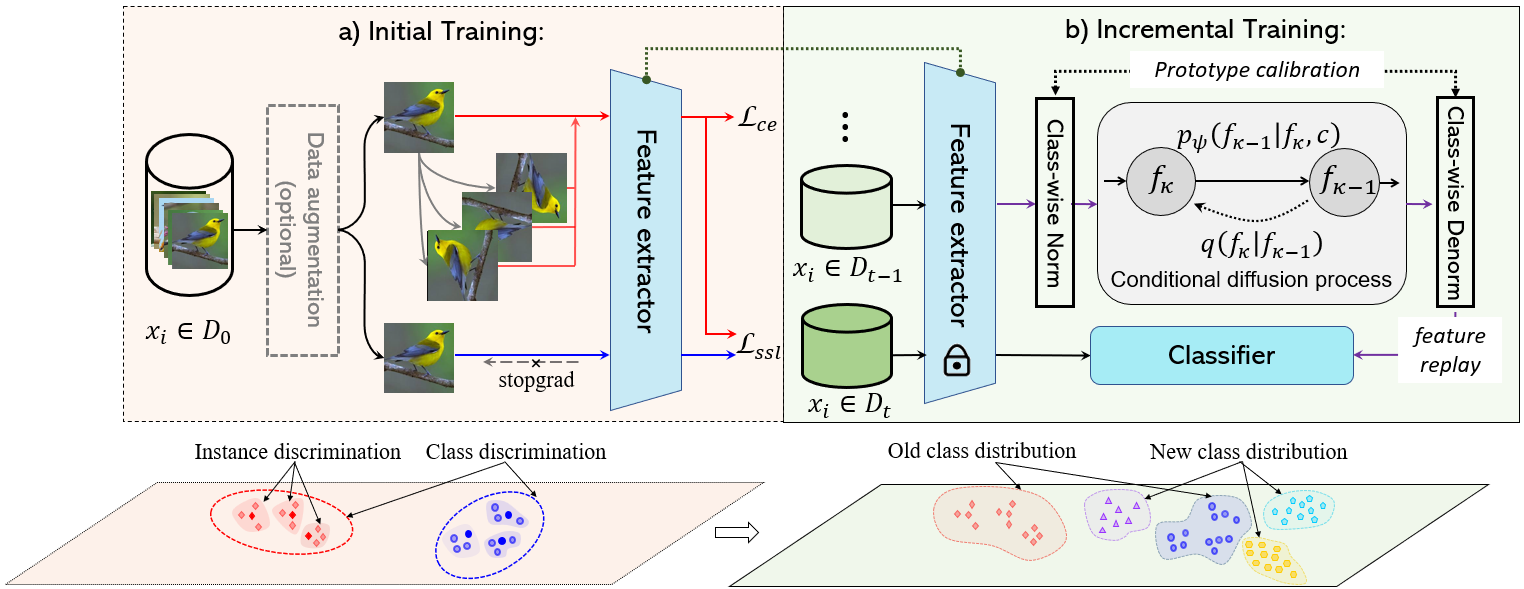}
  \caption{Illustration of DiffFR for NECIL. Classes of the initial task are augmented by rotation transformation, thus resulting the difference between the classifier in the initial training and that in the incremental training.}
  \label{fig:framework}
\end{figure*}

\textbf{Self-Supervised Learning.} 
Learning with self-supervision has proven to be effective to learn general representations. The recent studies begin with artificially designed pretext tasks, e.g., prediction rotations \cite{gidaris2018unsupervised}, patch permutation \cite{noroozi2016unsupervised} and image colorization \cite{larsson2016learning}. More recently, contrastive learning has achieved great success, and the core idea is to contrast positive pairs against negative pairs. In practice, this methodology benefits from a large number of negative samples \cite{wu2018unsupervised, tian2020contrastive, he2020momentum, chen2020simple}, resulting in the need for a memory bank \cite{he2020momentum} or a large batch size \cite{chen2020simple}. BYOL \cite{grill2020bootstrap} predicts the output of one view directly based on another view, obviating the necessity for a large batch size. SimSiam \cite{chen2021exploring} introduces the stop-gradient operation, which further eliminates the need for a momentum encoder, appearing simpler yet equally effective.

In CIL, some works explore improving the quality of the learned representations by SSL and prove its effectiveness. \cite{wu2021striking} attempts to learn class-agnostic knowledge and multi-perspective knowledge by SSL. PASS \cite{zhu2021prototype} employs a rotation-based proxy task to boost the performance, which can be viewed as a combination of supervised and self-supervised learning. However, pretext tasks rely on somewhat ad-hoc heuristics, which limits the generality of learned representations \cite{chen2020simple}. CaSSLe \cite{fini2022self} transforms the existing self-supervised loss function into distillation mechanisms, but it is entirely self-supervised, exhibiting relatively poor performance 
in CIL. In addition to supervised learning, we introduce self-supervised learning based on the simple and effective Siamese network, enhancing the generalization capability of the feature extractor.

\section{Methodology}
\textbf{Problem Statement.} In the paradigm of CIL, a unified model is required to be built from sequentially arrived training data to classify the test samples from all seen classes up so far. Specifically, the learning procedure consists of an initial phase and $T$ incremental ones. The incoming dataset $D_t$ at step $t$ has a form of $\{x_{i}^t,y_{i}^t\}_{i=1}^{n_t}$, where $x_{i}^t$ is the input image and $y_{i}^t$ is the attached label, $n_t$ is the number of labeled images for training. Correspondently, the class set at step $t$ is denoted as $C_t$. The class sets at different steps are mutually exclusive from each other, \textit{i.e.}, $C_t\cap C_{\tau}=\varnothing$, $t\neq \tau$. Let $\mathcal{C}_{1\sim t}=C_1\cup...\cup C_t $ be the cumulative label space and the classifier joint with feature extractor at step $t$ is expected to predict well on all classes in $\mathcal{C}_{1\sim t}$.

\textbf{Overview.} Fig.\ref{fig:framework} presents an overview of our proposed \textit{DiffFR}. In initial phase, the feature extractor $F_{\theta}$ is trained with initial data $D_0$. To obtain generalizable features and reduce representation confusion, we integrate label guidance into self-supervised learning to capture both class- and instance-level distribution. Stronger data augmentation can be further integrated for performance improvement. We also freeze the feature extractor $F_{\theta}$ for model stability, as in \cite{petit2023fetril, van2022three}. In $T$ incremental learning phases, we gradually update a unified classifier $G_{\phi}$ with the sequentially observed data $\{D_t\}_{t=1}^{T}$. Notably, we model a diffusion process to fit the feature representation and generate class-representative features of old classes. The generated features of old classes are merged with real features from new classes to update classifier $G_{\phi}$. Specially, prototype calibration is imposed to force the diffusion model focusing on the distribution shape. In \textit{DiffFR}, we memorize the distribution of feature representations with diffusion modeling, providing an effective solution for continual learning.

\subsection{Self-Supervision for Generalizable Features}

Considering that the feature extractor $F_{\theta}$ is trained on initial data $D_0$ and frozen in incremental phases, its generalization capability to unseen classes is crucial. To obtain the generalizable and transferable features, we propose to combine self-supervised learning and label guidance for the initial training of the feature extractor. Specifically, as in Fig.\ref{fig:framework}(a), an image $x_i \in D_0$ is augmented by two views. For one view, we additionally rotate the image $x_i$ by $90^\circ$, $180^\circ$, and $270^\circ$ and obtain the augmented data with new classes:
\begin{equation}
\setlength{\abovedisplayskip}{3pt}
    \tilde{x}_{i,j}=\mathtt{rotate}(x_i,j\times 90^\circ ), \tilde{y}_{i,j}=4y_i +j,  \label{eq:eq10}
\setlength{\belowdisplayshortskip}{3pt}
\end{equation}
where $j \in \{0,1,2,3\}$. It converts the original $N$-way classification into $4N$-way one. Cross-entropy loss is imposed on the output of extended classifier $\rho_c$, as:
\begin{equation}
\setlength{\abovedisplayskip}{3pt} \mathcal{L}_{ce}=\mathtt{CrossEntropy}\left(\rho_c \circ F_{\theta}(\tilde{x}_{i,j}),\tilde{y}_{i,j}\right) \label{eq:eq12}
\setlength{\belowdisplayshortskip}{3pt}
\end{equation}
For the other reference view, the data is directly from $x_i$ without rotation. We further perform Siamese-like self-supervised learning. The images of the two views pass through the feature extractor $F_{\theta}$ and an extra projector $\rho_s$. We then denote $\tilde{z}_{i,j}=\rho_s \circ F_{\theta}(\tilde{x}_{i,j})$, $z_{i}=\rho_s \circ F_{\theta}(x_{i})$. To ensure the feature of rotated images be similar to that of the non-rotated view, the self-supervised loss is formulated as:
\begin{equation}
\setlength{\abovedisplayskip}{3pt}
    \mathcal{L}_{ssl}=1-\mathtt{CosSimilarity}\left(\tilde{z}_{i,j}, \mathtt{stopgrad}(z_i)\right), \label{eq:eq14}
\setlength{\belowdisplayshortskip}{3pt}
\end{equation}
where $\mathtt{stopgrad}(\cdot)$ means gradient stopping in the reference view. We finally reach a total loss composed of the classification and self-supervised losses as:
\begin{equation}
\setlength{\abovedisplayskip}{3pt}
    \mathcal{L}_{all}=\mathcal{L}_{ce}+ \lambda \mathcal{L}_{ssl} \label{eq:eq13}
\setlength{\belowdisplayshortskip}{3pt}
\end{equation}
where $\lambda$ is the balancing weight. Typically $\lambda$ is set to 5.

The feature extractor $F_{\theta}$ is frozen after the initial training with enhanced classification and self-supervision. The initial classifier $G_{\phi}$ is learned based on frozen $F_{\theta}$ to classify the samples in $D_0$ and updated incrementally with newly arrived data $D_1, D_2, \cdots, D_t$. We train the feature extractor with consideration of both class- and instance-level representations, yielding features with reduced overlap, thus generalize and transfer well to unseen new classes.

\subsection{Diffusion-based Feature Replay}
We leverage generative models as the feature generator to memorize the knowledge of old classes. As for generative models, generative adversarial network (GAN) \cite{goodfellow2014generative, mirza2014conditional} suffers from training instability and label inconsistency in class-conditional generation \cite{ayub2020eec, ostapenko2019learning}, while variational autoencoder (VAE) \cite{kingma2013auto, sohn2015learning} generated data are relatively blurred \cite{wang2023comprehensive}. Considering the training stability and strong performance of diffusion models across different generative tasks, we adopt diffusion models as the feature generator to produce class-representative features exhibiting high similarity to real features.

We construct the diffusion process to generate features for replaying, thus $q(\mathbf{f}_k|\mathbf{f}_0)$ is defined as:
\begin{equation}
\setlength{\abovedisplayskip}{3pt} q(\mathbf{f}_k|\mathbf{f}_0):=\mathcal{N}(\mathbf{f}_k|\sqrt{\overline{\alpha}_k}\mathbf{f}_0,(1-\overline{\alpha}_k)I), \label{eq:eq1}
\setlength{\belowdisplayshortskip}{3pt}
\end{equation}
where $\overline{\alpha}_k:=\prod_{\kappa=0}^{k}\alpha_{\kappa}=\prod_{\kappa=0}^{k}(1-\beta_{\kappa})$ and $\beta_{\kappa}$ is the noise variance schedule. By Eq.(\ref{eq:eq1}), we directly obtain noisy samples $\mathbf{f}_k$ without step-by-step addition. In the denoising process for conditional generation, a network $u_{\psi}(\mathbf{f}_k,c,k)$ is trained to reverse $\mathbf{f}_0$ by iteratively predicting $\mathbf{f}_{k}$:
\begin{equation}
\setlength{\abovedisplayskip}{3pt}
    p_{\psi}(\mathbf{f}_{k-1}|\mathbf{f}_k,c):=\mathcal{N}(\mathbf{f}_{k-1};u_{\psi}(\mathbf{f}_k,c,k),\sigma_k^2I), \label{eq:eq3}
\setlength{\belowdisplayshortskip}{3pt}
\end{equation}
where $c$ is the conditional vector (\textit{i.e.,} class label) and $\sigma_k^2$ is the transition variance. The classifier-free guidance uses a guidance scale to control the trade-off between the quality and the diversity of the generated samples. Empirically we set the scale to 1 for a preference of diversity.

For prototype memorization and feature replay, we need to generate vectors with fixed dimension $d$ in incremental learning phases. Therefore, we treat a feature as a sequence with a single channel and the length equals $d$. We further design 1D U-Net \cite{ronneberger2015u} as the denoising network. Unlike the original U-Net in DDPM, we significantly simplify the network architecture to constrain the memory storage and achieve parameter efficiency. First, we remove the attention modules as it is deemed relatively less important. Second, except for the initial block, the regular convolutions are replaced with depthwise separable convolutions, to reduce computational cost with equal receptive fields. Third, we change the way of shortcut from concatenation to addition for feature reusing and further parameter reduction. We maintain the depth of the network and set the stride of the convolution layers as 4, achieving a larger downsampling rate. By utilizing these parameter reduction techniques, the diffusion model is quite small, while maintaining satisfactory performance.

\subsection{Prototype Calibration}

\begin{figure}[tbp]
  \centering
  \includegraphics[scale=0.3]{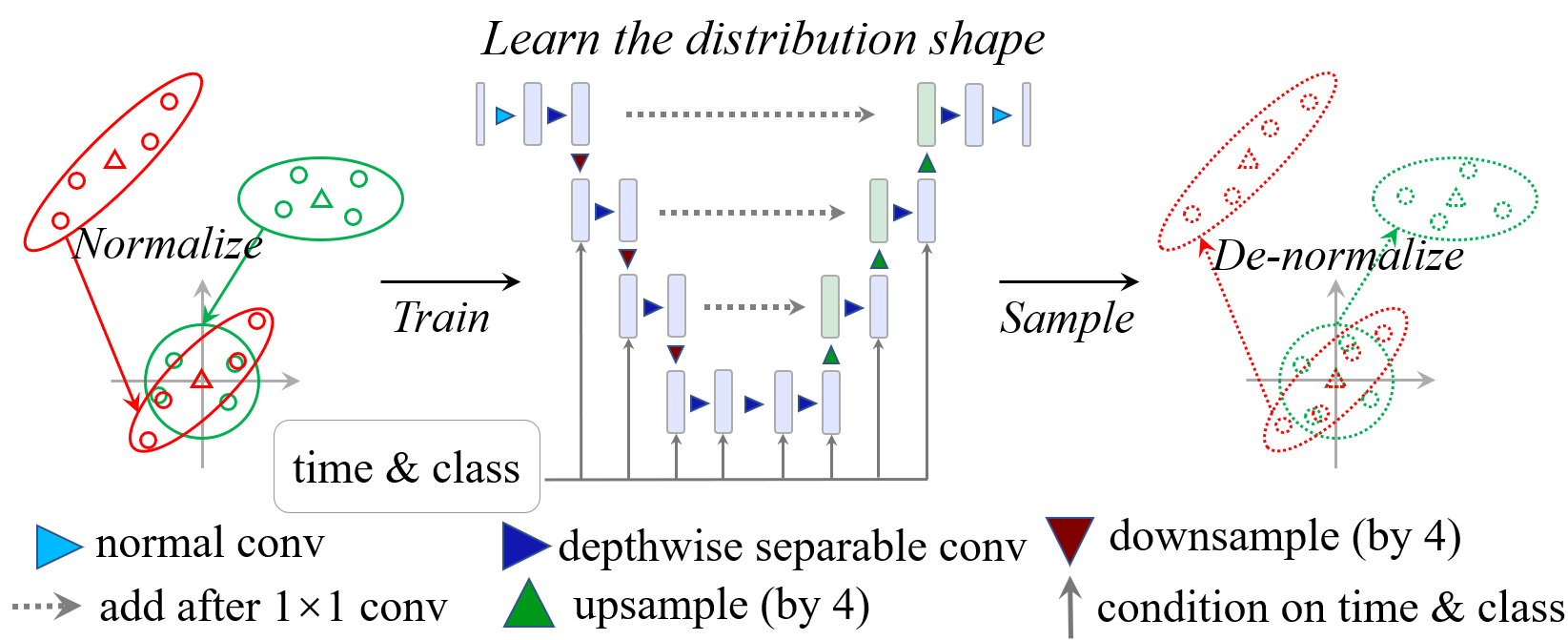}
  \caption{Illustration of prototype calibration. To learn the distribution of real features, we first normalize them by class. Subsequently, we use the diffusion model to learn the distribution of normalized features. Finally, denormalization by class is applied to samples to generate features with accurate prototypes.}
  \label{fig:network}
\end{figure}

In feature space, real features output from feature extractor of neural networks tend to show semantic clustering properties. Diffusion models provide a powerful way to learn the feature distribution of old classes for NECIL. However, the diffusion models are prone to capture the statistical values and overfit the location of clusters associated with different classes. That is, the diffusion models for feature generator would learn the statistics (\textit{i.e.}, means and variances) for shortcut solution, while ignores the shape of distribution with abundant information. To tackle this, we propose the prototype calibration which performs intra-class normalization and denormalization on generated features.

Specifically, as shown in Fig.\ref{fig:network}, we keep the mean vector and variance for real features of different classes along each dimension, denoted as $\mathbf{m}_c$ and $\mathbf{s}_c$, respectively. We then perform the "normalize by class" operation and obtain the centralized features. The feature distribution is simplified since the normalized features for each class have a mean of 0 and a variance of 1 along all dimensions. In diffusion process, the noise is gradually added on the normalized features. The denoised features are required to fit the normalized features as well. We further perform "denormalize by class" operation afterwards and feed the denormalized prototypes for feature replay. From the learning perspective, the prototype calibration ensures the diffusion models to concentrate on modeling the shape of feature clusters. From the feature representation perspective, the combination of recovered distribution from diffusion process and the statistical values provides more comprehensive information on feature representations. Thus the similarity between the generated features and real ones would be promoted. It would also help reduce the feature confusion among old and new classes, further benefit the class-incremental learning.

\section{Experiments}
In Sec.\ref{Experimental_Setup}, we elaborate on the experimental setup of NECIL and the detailed settings of domain incremental learning (DIL) can be found in the Appendix. In Sec.\ref{Quantitative_Results}, we present comprehensive experimental results and visualizations. In Sec.\ref{Comparative_Study}, we compare our approach with other diffusion-based methods for image replay. Beyond the challenging label-shift scenarios, we also showcase the performance of our method under domain-shift conditions. Finally, in Sec.\ref{Ablation_Study}, we perform detailed ablations to validate the effectiveness of diffusion-based feature replay and other components. In the Appendix, in addition to more detailed implementation and experimental results, we supplement the formulation of evaluation metrics, an analysis of similarity-based supervision, and the impact of the U-Net model structure.

\subsection{Experimental Setup}
\label{Experimental_Setup}

\noindent \textbf{Datasets and Settings.} We perform comprehensive experiments on three commonly used datasets: CIFAR-100 \cite{krizhevsky2009learning}, TinyImageNet \cite{le2015tiny} and ImageNet-Subset \cite{deng2009imagenet}. CIFAR-100 consists of 32$\times$32 pixel color images with 100 classes. It contains 50,000 images for training with 500 images per class, and 10,000 images for evaluation with 100 images per class. TinyImageNet contains 200 classes, with each class containing 500 images of 64 $\times$ 64 size. ImageNet-Subset is built by selecting 100 classes from ImageNet-Full (random seed 1993). Each dataset involves three incremental settings (5, 10, and 20 phases). For a fair comparison, the reordering and partitioning of classes are identical to \cite{zhu2021prototype}. Moreover, we adopt the same backbone network (\textit{i.e.}, ResNet-18 \cite{he2016deep}) and data augmentation strategies without specific noticing.

\noindent \textbf{Evaluation metric.} We report \textit{average incremental accuracy} \cite{rebuffi2017icarl} and \textit{average forgetting} \cite{chaudhry2018riemannian}. The average incremental accuracy $A_T$ is widely adopted as the main evaluation metric in CIL. It is computed as the average accuracy of all incremental phases, including the initial one. The average forgetting refers to the average difference between the final accuracy and the peak accuracy for each task throughout the incremental process, where a lower value is desirable.

\noindent\textbf{Implementation details.} 
Following CIL paradigm, our methodology encompasses three principal components: training the feature extractor, training the diffusion model, and incrementally updating the classifier. During the training of the feature extractor, we use SGD optimization, 100 epochs and a batch size of 32. The initial learning rate is 0.1 and decayed using a cosine annealing learning rate schedule \cite{loshchilov2016sgdr}. In the training of the diffusion model, we treat the feature vector as a sequence with a single channel, thus adopting an efficient 1-dimensional U-Net architecture. Class labels are one-hot mapped then projected into embedding vectors with 2-layer MLP. We train the model for 100k iterations with 20 diffusion steps and 20 inference steps. The batch size is 64 and learning rate is $8 \times 10^{-5}$. In the incremental training of the classifier, the per-class number of generated features generally equals that of real ones. We use SGD optimization with a batch size of 64 and a total of 20 epochs. The initial learning rate is 0.05 and is also decayed using a cosine annealing learning rate schedule.

\subsection{Quantitative Results}
\label{Quantitative_Results}

\begin{table*}[t]
  \label{sample-table}
  \centering
  \scalebox{0.78}{
  \begin{tabular}{lccccccccc}
    \toprule
    \multirow{2}{*}{\textbf{Methods}} &
    \multicolumn{3}{c}{CIFAR-100} & \multicolumn{3}{c}{TinyImageNet} & \multicolumn{3}{c}{ImageNet-Subset}  \\
    \cmidrule(r){2-4}
    \cmidrule(r){5-7}
    \cmidrule(r){8-10}
    &  $T$=5     & $T$=10     & $T$=20     & $T$=5     & $T$=10     & $T$=20     & $T$=5     & $T$=10     & $T$=20 \\
    \midrule
    EWC \cite{kirkpatrick2017overcoming} & 24.5  & 21.2  & 15.9  &  18.8  & 15.8  & 12.4  & -  & 20.4  & -  \\
    LwF-MC \cite{rebuffi2017icarl}  & 45.9 & 27.4 & 20.1 &  29.1   & 23.1   & 17.4  & -  & 31.2  & -  \\
    DeeSIL \cite{belouadah2018deesil} & 60.0   & 50.6   & 38.1  & 49.8   & 43.9   & 34.1  &  67.9   & 60.1   & 50.5  \\
    LUCIR \cite{hou2019learning} & 51.2   & 41.1   & 25.2  & 41.7   & 28.1   & 18.9  &  56.8   & 41.4   & 28.5  \\
    MUC \cite{liu2020more}  & 49.4   & 30.2   & 21.3  &  32.6   & 26.6   & 21.9  & -  & 35.1  & -  \\
    SDC \cite{yu2020semantic}  & 56.8   & 57.0   & 58.9  & -  & -  & -  & -  & 61.2  & -  \\
    ABD \cite{smith2021always}  & 63.8   & 62.5   & 57.4  & -  & -  & -  & -  & -  & -  \\
    PASS \cite{zhu2021prototype}  & 63.5   & 61.8   & 58.1  &  49.6   & 47.3   & 42.1  &  64.4   & 61.8   & 51.3  \\
    IL2A \cite{zhu2021class}  & 66.0   & 60.3   & 57.9  &  47.3   & 44.7   & 40.0  & -  & -  & -  \\
    SSRE \cite{zhu2022self}  & 65.9   & 65.0   & 61.7  &  50.4   & 48.9   & 48.2  & - & 67.7  & -  \\
    FeTrIL \cite{petit2023fetril} & 66.3   & 65.2   & 61.5  &  \underline{54.8}  & \underline{53.1}    & \underline{52.2}  &  \underline{72.2}   & \underline{71.2}   & \underline{67.1}  \\
    SOPE \cite{zhu2023self}  & 66.6   & 65.8   & 61.8  &  53.7   & 52.9   & 51.9  & - & 69.2  & -  \\
    PRAKA \cite{shi2023prototype} & \underline{70.0}   & \underline{68.9}   & \underline{65.9}  &  53.3   & 52.6   & 49.8  & - & 69.0  & -  \\
    \midrule
    DiffFR (ours)      & 
    \textbf{72.2}$^{+2.2}$  & \textbf{71.9}$^{+3.0}$  & \textbf{70.7}$^{+4.8}$  & \textbf{56.4}$^{+1.6}$  & \textbf{55.9}$^{+2.8}$  & \textbf{55.8}$^{+3.6}$  & \textbf{73.7}$^{+1.5}$  & \textbf{73.4}$^{+2.2}$  & \textbf{72.1}$^{+5.0}$  \\
    DiffFR* (ours)       &\textbf{72.8}$^{+2.8}$  & \textbf{72.7}$^{+3.8}$  & \textbf{72.1}$^{+6.2}$  & \textbf{62.7}$^{+7.9}$  & \textbf{62.5}$^{+9.4}$  & \textbf{62.5}$^{+10.3}$  & \textbf{76.1}$^{+3.9}$  & \textbf{76.0}$^{+4.8}$  & \textbf{73.9}$^{+6.8}$  \\
    \bottomrule
  \end{tabular}}
  \caption{The average incremental accuracy $(\uparrow)$ in NECIL with different incremental phases on CIFAR-100, TinyImageNet, and ImageNet-Subset. We use the superscript to report the improvement over existing state-of-the-art results. ”-” means that results are unavailable. \textit{*} means enhanced data augmentation for initial training. The best and past SOTA results are in \textbf{bold} and \underline{underlined}.}
  \label{table:table1}
\end{table*}

\begin{table}[h]
\centering
\scalebox{0.85}{
\begin{tabular}{lcccccc}
\toprule
\multirow{2}{*}{Methods}   & \multicolumn{3}{c}{CIFAR-100} & \multicolumn{3}{c}{TinyImageNet}       \\
\cmidrule(r){2-4}
\cmidrule(r){5-7}
& $T$=5       & $T$=10       & $T$=20 & $T$=5       & $T$=10       & $T$=20       \\
\midrule
LwF\_MC     & 44.23 & 50.47 & 55.46 & 54.26 & 54.37 & 63.54           \\
MUC     & 40.28 & 47.56 & 52.65 & 51.46 & 50.21 & 58.00           \\
PASS     & 25.20 & 30.25 & 30.61 & 18.04 & 23.11 & 30.55           \\
SSRE     & 18.37 & 19.48 & 19.00 & 9.17 & 14.06 & 14.20           \\
SOPE     & 6.50   & \textbf{3.30}   & 9.14   & -   & -   & -           \\
PRAKA     & 7.18 & 6.42 & 10.30 & 4.57 & 5.10 & 9.05           \\
\midrule
DiffFR (ours)    & \textbf{4.31} & 6.02 & \textbf{7.90} & \textbf{2.27} & \textbf{3.66} & \textbf{5.03}           \\
\bottomrule
\end{tabular}}
\caption{Results of average forgetting $(\downarrow)$ on CIFAR-100 and TinyImageNet.}
\label{table:forgetting}
\end{table}

We compare the proposed DiffFR with a wide range of NECIL methods, including EWC \cite{kirkpatrick2017overcoming}, LwF-MC \cite{rebuffi2017icarl}, DeeSIL \cite{belouadah2018deesil}, LUCIR \cite{hou2019learning}, MUC \cite{liu2020more}, SDC \cite{yu2020semantic}, ABD \cite{smith2021always}, PASS \cite{zhu2021prototype}, IL2A \cite{zhu2021class}, SSRE \cite{zhu2022self}, FeTrIL \cite{petit2023fetril}, SOPE \cite{zhu2023self} , PRAKA \cite{shi2023prototype}. The similarity-based self-supervision in our method requires more enhanced data augmentation such as Cutout \cite{deVries2017cutout} to be more effective. For a fair comparison, we provide results for DiffFR under both the same and enhanced data augmentation conditions. Additionally, in supplementary material, we reproduce the recent methods (ABD, PASS, IL2A, SSRE and FeTrIL) under enhanced data augmentation to illustrate the more pronounced superiority of DiffFR and the effect of the self-supervision.

\begin{figure*}[t]
  \centering
  \includegraphics[scale=0.086]{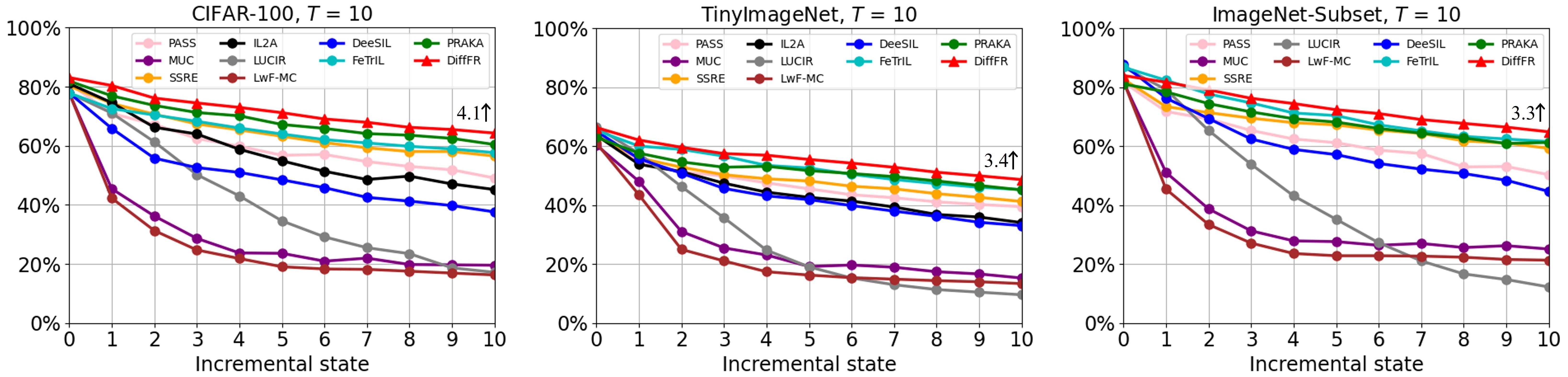}
  \caption{Evolution of top-1 accuracy on three datasets with T = 10 phases.} 
  \label{fig:curve_1}
\end{figure*}

\begin{figure*}[h]
  \centering
  \includegraphics[scale=0.135]{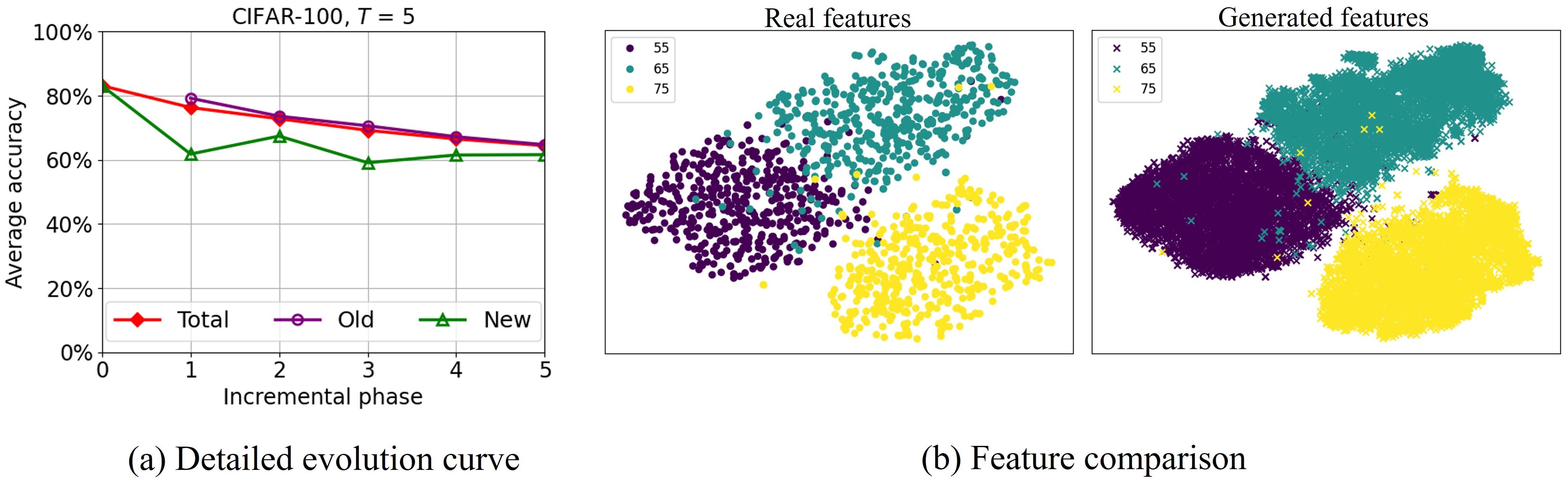}
  \caption{(a) Average accuracy of total/old/new classes at each phase. For better visualization, we use the setting of CIFAR-100,$T=5$ as an example. (b) Visualization of real and generated features on CIFAR-100 (10 phases). Each class has 500 real features, and we generate 5000 features to facilitate more accurate comparison between the distribution shape of real features and generated features.} 
  \label{fig:vis_2}
\end{figure*}

\begin{table*}[t]
\begin{floatrow}
\capbtabbox{
\scalebox{0.85}{
\begin{tabular}{l|ccc}
\toprule
Method & Final Acc. $(\uparrow)$    & Params $(\downarrow)$    & Steps $(\downarrow)$  \\
\midrule
DDGR \cite{gao2023ddgr} &56.3 &130 M  &4000 \\
 SSDR \cite{jodelet2023class} &50.9 & 860 M  &50  \\
 \midrule
 DiffFR (ours) &\textbf{61.0} &\textbf{10 M}  &\textbf{20}\\
\bottomrule
\end{tabular}}
}{
 \caption{Comparison with other diffusion-based methods in incremental learning.}
 \label{tab:DDGR}
 \small
}

\capbtabbox{
\scalebox{0.85}{
\begin{tabular}{l|ccc}
\toprule
Method & $T$=5     & $T$=10     & $T$=20  \\
\midrule
GAN \cite{mirza2014conditional} &67.8 &67.1  &63.2 \\
 VAE \cite{sohn2015learning} &68.1 &67.5  &63.2  \\
 \midrule
 Diffusion (ours) &\textbf{72.2} &\textbf{71.9}  &\textbf{70.7}\\
\bottomrule
\end{tabular}}
}{
 \caption{Comparison with other generative models.}
 \label{table:GAN}
}
\end{floatrow}
\end{table*}

We conduct different incremental settings (5, 10 and 20 phases) for all three datasets and results from Tab.\ref{table:table1} show that DiffFR outperforms existing methods by a large margin. Take the results of 10 phases as an example, DiffFR outperforms the state-of-the-art (SOTA) results by 3.0\% on CIFAR-100, 2.8\% on TinyImageNet, and 2.2\% on ImageNet-Subset, respectively. In Fig.\ref{fig:curve_1}, we present comprehensive accuracy curves on three benchmarks with $T$=10, offering a more detailed view of the accuracy evolution during the CIL process. With similar or even lower accuracy (see ImageNet-Subset with $T$=10) in the initial stage, DiffFR exhibits higher accuracy in subsequent stages. The accuracy of the final stage has improved by 4.1\%, 3.4\%, and 3.3\% compared to the SOTA across the three benchmarks. Additionally, we provide the results of average forgetting in Tab.\ref{table:forgetting}. Apart from lagging behind SOPE on CIFAR-100 (10 phases), we consistently achieve the best performance in other scenarios.



It can be seen from Tab.\ref{table:table1} and \ref{table:forgetting} that DiffFR exhibits significant performance gains over existing methods. Compared to FeTrIL which also employs a frozen feature extractor and classifier updating strategy, DiffFR achieves the average improvement of 7.3\%, 2.7\% and 2.9\% on CIFAR-100, TinyImageNet, ImageNet-Subset datasets. Compared to PASS which also involves self-supervised learning, DiffFR achieves more significant gains, which validates the importance of Siamese-based self-supervised learning and diffusion-based feature replay.

It is noteworthy that, as the incremental phase $T$ increases, DiffFR experiences minimal performance degradation. For instance, when transitioning from CIFAR-100, $T$=5 to $T$=20, most existing methods exhibit an performance (average accuracy) drop of 4\%$\sim$8\%, whereas DiffFR only experiences a 1.5\% performance decrease. This leads to a greater advantage for DiffFR at larger $T$ values. Specifically, DiffFR outperforms SOTA methods by an average of 1.8\% on three datasets with $T$=5 and by an average of 4.5\% with $T$=20.

Besides, we illustrate the detailed accuracy curve for old and new classes in Fig.\ref{fig:vis_2}(a). We follow the common setting in incremental learning to freeze the feature extractor \cite{van2022three}. It can be seen from the figure that the frozen feature extractor generalizes well, thus yielding subtle accuracy changes for new classes. 


\noindent \textbf{Visualization for generated features.} 
To validate the efficacy of diffusion-based feature replay, we visualize the 2D embeddings of feature vectors with t-SNE \cite{maaten2008tsne} in Fig.\ref{fig:vis_2}(b). The replayed features exhibit a high degree of similarity to the real features. In detail, replayed features and real features not only share similar means but also align in terms of distribution shape, which is not achievable through the prototype-based simple rules approach for feature generation.

\subsection{Comparative Study}
\label{Comparative_Study}

\noindent \textbf{Comparison with diffusion-based CIL methods.} Existing methods such as DDGR \cite{gao2023ddgr} and SSDR \cite{jodelet2023class} adopt diffusion models to generate images for replaying. In contrast, the proposed DiffFR contructs the diffusion process for feature replay. Since diffusion models for high-dimensional image generation is challenged by quality, DDGR mainly tackles the low-quality fuzzy generative images, while SSDR employs pre-trained stable diffusion. In contrast, DiffFR focuses on feature replay, with the core idea that feature is the essence of memorization in continual learning. Comparative study under the same setting with SSDR on CIFAR-100 (T=5) is presented in Tab.\ref{tab:DDGR}.  Generally, DDGR surpasses SSDR on accuracy, with the requirements of much more training steps. Compared to DDGR, the proposed DiffFR achieves 4.7\% final accuracy improvement with 8\% parameters and 0.5\%  diffusion steps, validates both the effectiveness and efficiency of our diffusion-based feature replay. 

\noindent \textbf{Comparison of diffusion model over other generative models.}
We replace diffusion with GAN \cite{goodfellow2014generative, mirza2014conditional} and VAE \cite{kingma2013auto, sohn2015learning}, to study the effectiveness of diffusion models for generative feature replay. The achieved accuracy in CIFAR-100 under $T=10$ setting is 67.1\% and 67.5\% for GAN and VAE based generative feature replay. In contrast, the feature replay with diffusion model obtains 71.9\% under the same setting. It is reasonable that diffusion is better in modeling distribution shapes, leading to better memorization and replay.


\noindent \textbf{Comparison beyond label-shift scenarios.}
To validate the adaptability of the DiffFR in other increment learning tasks, we conduct experiments on domain incremental learning (DIL) with domain-shift. As shown in Tab.\ref{tab:DIL}, our method achieves impressive results, significantly outperforming SOTA methods across all settings. Specifically, DiffFR achieves remarkable performance on Seq-CORe50, with 86.5\% average accuracy and 10.6\% forgetting, compared to the SOTA result which records 82.1\% average accuracy and 19.6\% forgetting. This is reasonable, as the domain-shift scenario is relatively less challenging due to the smaller differences between domains compared to the differences between labels.

\begin{table*}[t]
\begin{floatrow}

\capbtabbox{
\scalebox{0.7}{
\begin{tabular}{c|cccc}
\toprule
    \multirow{2}{*}{Methods}     & \multicolumn{2}{c}{R-MNIST}    & \multicolumn{2}{c}{Seq-CORe50} \\
    \cmidrule(r){2-3}
    \cmidrule(r){4-5}
    
     & Avg. Acc $(\uparrow)$    & Fgt. $(\downarrow)$   & Avg. Acc $(\uparrow)$    & Fgt. $(\downarrow)$ \\
    
    \midrule
    ER \cite{riemer2018learning} &76.8 &20.7 &66.6 &32.8  \\
    DER++  \cite{buzzega2020dark} &84.3 &13.7 &78.6 &21.9  \\
    CLS-ER  \cite{arani2022learning} &81.8  &15.5 &- &-  \\
    ESM-ER  \cite{sarfraz2023error} &82.2  &16.2 &52.8 &25.4  \\
    UDIL  \cite{shi2024unified} &86.6  &8.5 &82.1 &19.6  \\
    \midrule
    DiffFR (ours)  &\textbf{89.4}  &\textbf{5.7} &\textbf{86.5} &\textbf{10.6}  \\
    \bottomrule
\end{tabular}}
}{
 \caption{Performances (\%) in DIL. We evaluate different methods on R-MNIST \cite{mnist} and Seq-CORe50 \cite{lomonaco2017core50, lomonaco2020rehearsal} with two metrics, Average Accuracy (Avg. Acc) and Forgetting (Fgt.). $(\uparrow)$ and $(\downarrow)$ mean higher and lower numbers are better, respectively. "-" denotes out-of-memory (OOM) error when running the experiments. Our method does not save any exemplars, whereas all other methods keep a memory size of 800 for replay.}
 \label{tab:DIL}
 \small
}

\capbtabbox{
\scalebox{0.7}{
\begin{tabular}{ccccccccc}
\toprule
\multicolumn{5}{c}{Components} & \multicolumn{3}{c}{CIFAR-100} \\
\cmidrule(r){1-5}
\cmidrule(r){6-8}
DFR       & PC & LA      & EA      & SIM      & $T$=5     & $T$=10     & $T$=20     \\
\midrule
          &         & &         &          & 23.1        & 13.1         & 8.0         \\
\checkmark          & &         &         &          & 68.3        & 67.8         & 66.2         \\
\checkmark          & \checkmark        &  &         &          & 69.0        & 68.7         & 67.3         \\
\checkmark          & \checkmark        & \checkmark         &  &         & 70.2        & 69.7         & 68.5 \\
\midrule
\checkmark          & \checkmark         & \checkmark       & \checkmark  &       & 71.8        & 71.6         & 70.6 \\
\checkmark          & \checkmark         & \checkmark       & \checkmark  & \checkmark      & \textbf{72.8}        & \textbf{72.7}         & \textbf{72.1} \\
\bottomrule
\end{tabular}}
}{
 \caption{Effects of core components. DFR means diffusion-based feature replay. PC means prototype calibration. LA means rotation-based label augmentation. EA means enhanced data augmentation and SIM means similarity-based self-supervision of our method.}
 \label{table:ablation}
}

\end{floatrow}
\end{table*}

\subsection{Ablation Study}
\label{Ablation_Study}

\textbf{Core components.} Tab.\ref{table:ablation} summarizes the results of our ablative experiments on CIFAR-100. Diffusion-based feature replay is the foundation of our approach. Without replaying features, the feature-level catastrophic forgetting occurs. In fact, we can utilize a simple approach, masked softmax cross-entropy \cite{caccia2021new}, to mitigate the classifier drift for old classes, which achieves results of 59.3\%, 58.7\%, and 55.1\% on CIFAR-100, respectively. Since the classifier boundary changes along with new class features, it would be better to update the classifiers synchronously, thus further highlighting the importance of feature replay.

Our proposed training method, which combines class- and instance-level representations effectively, results in an average accuracy improvement of 3.1\%, owing to the promotion of feature extractor generalization. In addition to these pivotal components, prototype calibration reduces the complexity of feature distributions, simplifying the task from learning the distribution to learning the distribution shape, consequently improving performance by average 0.9\%.

\noindent \textbf{Effects of diffusion steps and sampling steps.} A primary drawback of the diffusion model is the slow sampling speed due to the necessity for iterative denoising. In most experiments, both the diffusion step and sampling step are set to 20 during training and testing, without employing DDIM sampling. Altering either of them yields more results presented in Tab.\ref{table:diffusionsteps}. It can be found that 10$\sim$20 diffusion steps is sufficient for learning the feature distribution, which further corroborates the much lower complexity of the feature distribution compared to the image distribution. The low-dimensional nature of the features and the small sampling step contribute to the fast training and inference of DiffFR.

\begin{table}[t]
\centering
\scalebox{0.8}{
\begin{tabular}{p{1.1cm}<{\centering}|p{1.1cm}<{\centering}p{1.1cm}<{\centering}p{1.1cm}<{\centering}|p{1.1cm}<{\centering}p{1.1cm}<{\centering}p{1.1cm}<{\centering}}
\toprule
\multirow{2}{*}{Steps}   & \multicolumn{3}{c|}{\small{Diffusion training}}  & \multicolumn{3}{c}{\small{Sampling w/o re-training}}  \\
& $T$=5       & $T$=10       & $T$=20    & $T$=5       & $T$=10       & $T$=20    \\
\midrule
10   & \textbf{72.8}          & \textbf{72.7}           & \textbf{72.1}    & 72.7          & 72.7           & 72.1    \\
5    & 72.4          & 72.3           & 71.5    & 72.6          & 72.4           & 71.8         \\
2    & 71.9          & 71.7           & 70.8    & 72.4          & 72.3           & 71.8         \\
1    & 70.8          & 70.4           & 69.3    & 68.9          & 68.6           & 68.0        \\
\bottomrule
\end{tabular}}
\caption{Ablation study on different diffusion steps and sampling steps on CIFAR-100.}
\label{table:diffusionsteps}
\end{table}

\noindent \textbf{Storage overhead and runtime cost.} The additional storage overhead of our method primarily arises from the independent feature generators at each stage. The features are low-dimensional and tend to cluster together in the feature space. Moreover, we employ an efficient U-Net with reduced parameters. Therefore, the additional storage overhead in our method is acceptable. Taking ImageNet-Subset with 5 phases and 10 phases as examples, we require an additional 9.5M and 15.5M parameters, respectively. In contrast, distillation-based methods necessitate storing the old model, incurring an additional parameter requirement of at least 11.4M, and directly storing features would result in a memory need of 60M. Therefore, as analyzed in the introduction, using a separate feature generator at each stage is acceptable, and over time, multiple feature generators can be distilled into one to further reduce storage overhead. As for runtime cost, no extra time cost is needed in testing. The training runtime at each incremental phase of DiffFR(0.3h) is on par with FeTrIL(0.1h), much less than PRAKA(2.3h). Comparisons are based on TinyImageNet and a 3090 GPU.

\section{Conclusion}

In this work, we propose DiffFR to counteract the catastrophic forgetting in NECIL. We advance the NECIL from two aspects: (i) improving the generalization of the feature extractor, (ii) increasing the similarity between generated features and real features for effective replay. In particular, we introduce similarity-based self-supervised learning to mitigate the potential distribution overlap for different categories, so that the feature extractor can generalize and transfer well even for unseen classes. More importantly, we propose the diffusion model with prototype calibration to replay features for exemplar-free class-incremental learning. The diffusion model has been proven effective to generate features highly similar to real ones, which is suitable to memorize the feature scatters for replay. Extensive experiments demonstrate that the proposed DiffFR surpasses existing NECIL methods by a significant margin. 

\printbibliography


\appendix

\section{Supplementary Material}

This supplementary material mainly provides more implementation and evaluation details that have not been enclosed in the main paper due to page limit. In Sec.\ref{sec:1}, we provide more details about the implementation and results in the NECIL (non-exemplar class-incremental learning) experiments. Subsequently, we further analyze the effectiveness of our DiffFR in Sec.\ref{sec:2}, including more ablation experiments and visualizations of similarity-based self-supervision in Section \ref{sec:2_1}, visualizations of features from different methods in Section \ref{sec:2_2}, and the performance of different methods under enhanced data augmentation in Section \ref{sec:2_3}. Following this, we present the experimental setups and details of DiffFR in the domain-incremental learning (DIL) in Sec.\ref{sec:3}. Finally, we discuss future work in Sec.\ref{sec:4}, suggesting that applying similarity-based self-supervision and diffusion-based feature replay to the distillation-based framework is a promising avenue for exploration.

\section{More Details for Class-Incremental Learning (CIL)}
\label{sec:1}

\subsection{Formulation of Evaluation Metrics}
In the main paper, we report the average incremental accuracy and average forgetting. The detailed formulation of these metrics are presented in this supplementary material for reference. The average incremental accuracy $A_T$ calculates the mean accuracy over $T$ phases, including the initial one. At each phase $t$, the accuracy $a_t$ is computed based on all seen classes. Therefore, the average incremental accuracy $A_T$ can be expressed as:
\begin{equation}
\setlength{\abovedisplayskip}{3pt}
    A_T=\frac{1}{T+1}\sum\limits_{t=0}^{T}a_t.
\setlength{\belowdisplayshortskip}{3pt}
\end{equation}

Besides of average incremental accuracy $A_T$, the average forgetting is another important metric for CIL. 
It reflects the mean forgetting throughout the entire incremental process, directly measuring the capability to resist \textit{catastrophic forgetting}. The forgetting at phase $t (t>0)$ is calculated as:
\begin{equation}
\setlength{\abovedisplayskip}{3pt}
    F_t=\frac{1}{t}\sum\limits_{j=0}^{t-1}f_j^{t},
\setlength{\belowdisplayshortskip}{3pt}
\end{equation}
where the forgetting $f_j^{t}$ of task $j$ is defined as the difference between the peak accuracy and the current accuracy, expressed as:
\begin{equation}
\setlength{\abovedisplayskip}{3pt}
    f_j^{t}=\mathop{\max}_{i \in \{j,...,t-1\}} \left(a_{i,j}-a_{t,j}\right).
\setlength{\belowdisplayshortskip}{3pt}
\end{equation}
The average forgetting of the entire incremental learning process can be further formulated as follows:
\begin{equation}
\setlength{\abovedisplayskip}{3pt}
    F_T=\frac{1}{T}\sum\limits_{t=1}^{T}F_t.
\setlength{\belowdisplayshortskip}{3pt}
\end{equation}

\subsection{More Implementation Details}
In the training of the feature extractor, we employ a large learning rate of 0.1 and a small batch size of 32 to promote the generalization. We use the SGD optimizer with a cosine annealing learning rate schedule and initial training consists of 100 epochs. On the supervised branch, owing to rotation-based label augmentation, the output dimension of the classifier is $4C_0$, where $C_0$ denotes the number of classes in the initial phase. On the self-supervised branch, we refer to the feature extractor and the subsequent 2-layer MLP network (projector) as the encoder. Two views of the same image pass through the encoder, resulting in two features. Subsequently, one of them is used to predict the other through a predictor, which is also a 2-layer MLP network. To prevent model collapse, we use stop-gradient during the training. After the initial training, we retain the feature extractor with strong generalization and discard the remaining components. The feature extractor simplifies complex image streams into simple feature streams, allowing for replay at the feature level efficiently.

In the training of diffusion models, we utilize the Exponential Moving Average (EMA) strategy with a smoothing weight of 0.995. In the training of the classifier, we discard the classifier with an output dimension of $4C_0$ in the initial training. Then we freeze the feature extractor and retrain a new classifier with an output dimension of $C_0$ on the initial data, followed by incremental updates. In each update, we only train a single linear layer with minimal computational cost. Details on U-Net are in Sec.\ref{detail_U-Net}, while other experimental details of less importance will be available in our code.

\subsection{More Comparison Details over Existing Diffusion based Methods}
There are mainly two methods which also use diffusion in class-incremental learning, such as DDGR and SSDR. We have provided quantitative comparison in the main paper. The original setting in DDGR paper partitions data in a way different from existing methods (including SSDR). Considering that SSDR which involves pre-trained stable diffusion has not been open-sourced to date, we follow the setting of data partitioning in SSDR paper and evaluate the original codes of DDGR and our DiffFR under the same setting for fair comparisons. Specifically, experiments are conducted on CIFAR-100, with a 32-layer ResNet serving as the backbone. The class order is shuffled by NumPy using the random seed 1993. The initial phase contains half of the classes, with the remaining half evenly distributed across $5$ incremental phases. Training images are normalized, randomly horizontally flipped, and cropped, without other data augmentation techniques.

For SSDR, since its performance under the non-exemplar setting is not provided, we report its best performance under the condition of 5 exemplars/class, which is from its supplementary material. Due to not being specifically designed for NECIL, its performance is relatively worse compared to DDGR and DiffFR. 

\begin{figure*}[tbp]
  \centering
  \includegraphics[scale=0.19]{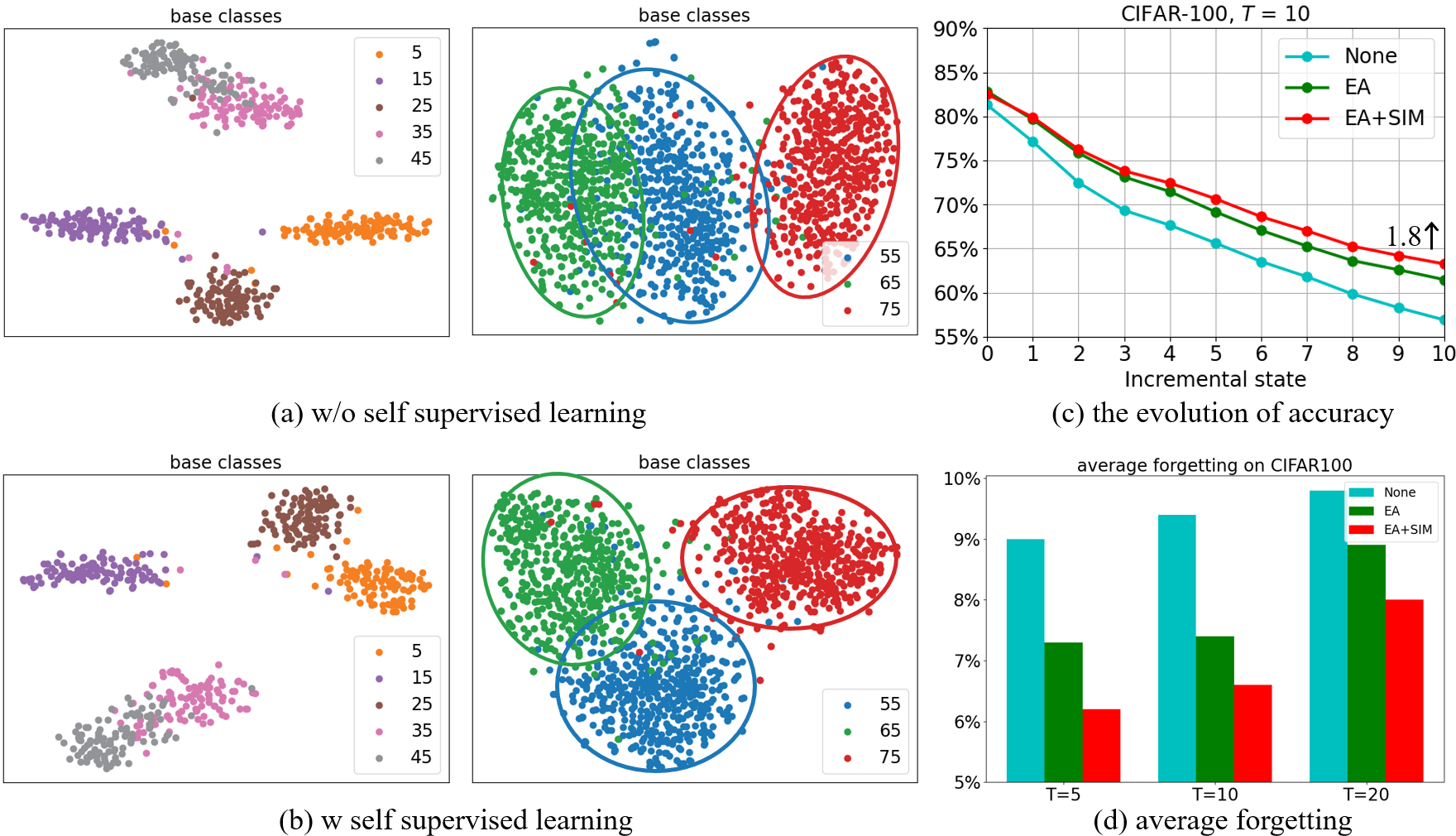}
  \caption{(a-b) Visual comparison of features extracted by the feature extractor with different training methods. The features of base classes exhibit a small difference, while the features of novel classes experience a reduction in overlap. (c) Comparison of the evolution of top-1 accuracy on CIFAR100 T=10. (d) Quantitative comparison of average forgetting on CIFAR100 across 5, 10 and 20 phases.}
  \label{fig:u_net}
  \vspace{-0.2cm}
\end{figure*}

\begin{figure*}[tbp]
  \centering
  \includegraphics[scale=0.26]{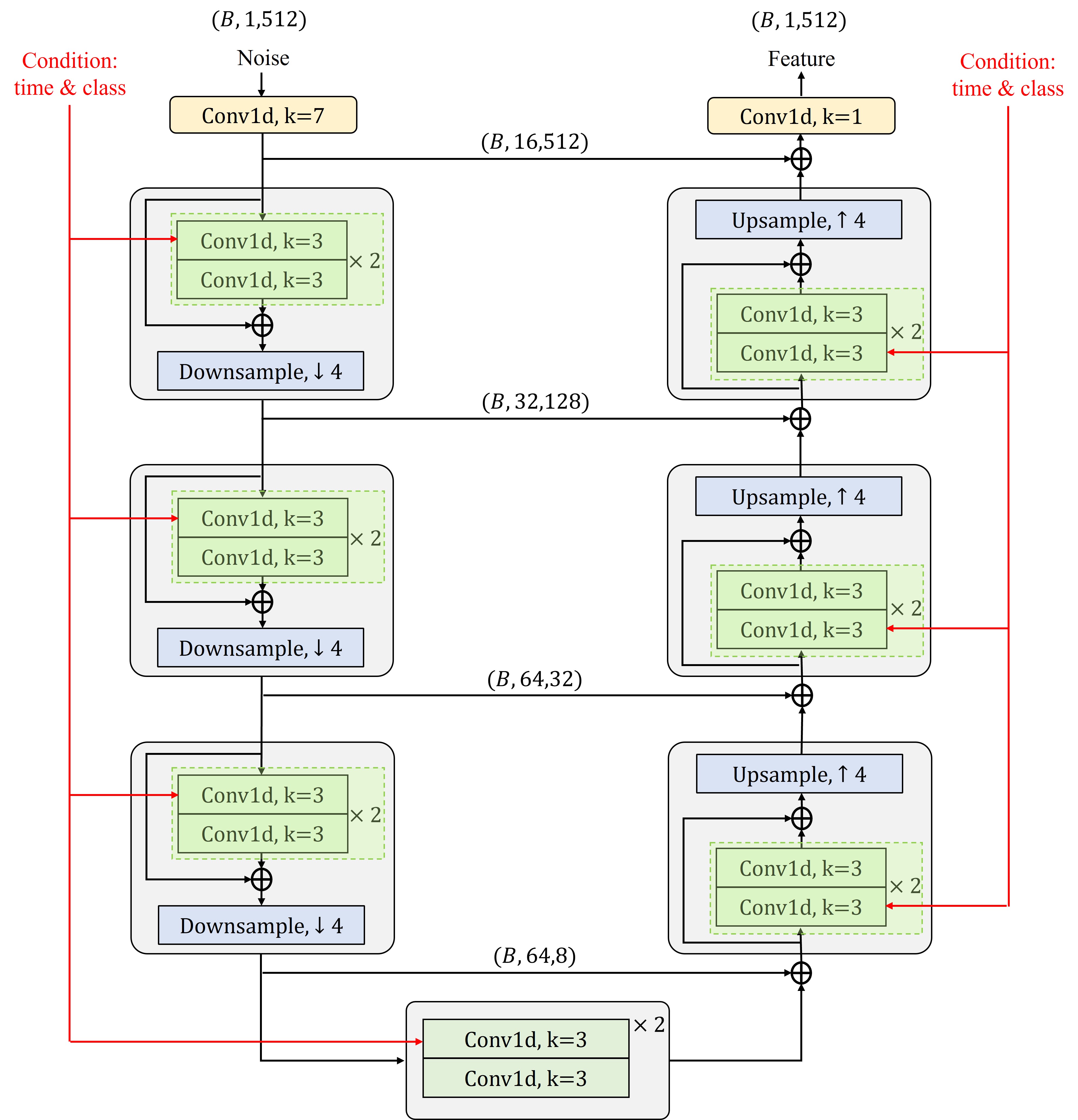}
  \caption{The U-Net architecture in our DiffFR. The yellow and green modules denote conventional convolution and depthwise separable convolution, respectively.}
  \label{fig:ssl_effect}
  \vspace{-0.2cm}
\end{figure*}

\section{More Experimental Analysis in CIL}
\label{sec:2}

\subsection{Effect of Similarity-based Self-supervision} 
\label{sec:2_1}
Given the proven efficacy of rotation-based label augmentation (LA) in PASS and PRAKA, we elaborate on the effectiveness of similarity-based self-supervision. Note that all experiments below exclude label augmentation (LA). We designate the classes in the initial stage as base classes, and those in subsequent stages as novel classes. In the comparisons depicted in Fig.\ref{fig:ssl_effect}(a-b), self-supervised learning has a negligible impact on base classes. However, as it promotes the learning of non-task-specific features, the generalization of the feature extractor is enhanced. This leads to a reduction in overlap among features of new classes, consequently improving overall performance. These analyses are supported by Fig.\ref{fig:ssl_effect}(c). With the incorporation of similarity-based supervision, the classification accuracy in the initial stage remains relatively unchanged. However, a clear advantage emerges in subsequent stages, resulting in a 1.8\% improvement in accuracy of the final stage. Additionally, in Fig.\ref{fig:ssl_effect}(d), we present the results of average forgetting across three settings on CIFAR-100, demonstrating the effective reduction of forgetting achieved by our method.

\subsection{Detailed Design of U-Net Structure and Its Effect}
\label{detail_U-Net}
The 1-dimensional U-Net used for feature replay is composed of down-sampling, middle, and up-sampling modules. They consist of 3, 1, and 3 blocks, respectively, with each block typically containing 4 depthwise separable convolutions. In the down-sampling, for the first 4 blocks, features are down-sampled by a factor of 4, and the feature dimensions in the 5 blocks are generally 32, 32, 64, 64, 128. If replaying initial data with more classes, the feature dimensions are doubled. The up-sampling module is symmetric to the down-sampling module, and long skip connections are present in corresponding modules to facilitate learning. 

Our 1-D U-Net aims to reduce parameters while keeping the performance of incremental learning: removing attention(Avg. Acc.:0.3\%$\downarrow$, params:4M$\downarrow$), regular convolutions$\rightarrow$separable ones(Avg. Acc.:0.5\%$\downarrow$, params:6M$\downarrow$), concatenate$\rightarrow$add (Avg. Acc.:0.2\%$\uparrow$, params:1M$\downarrow$). Downsampling rate from 2 to 4 (i.e., feature length from 512$\rightarrow$256$\rightarrow$128 to 512$\rightarrow$128$\rightarrow$32) improves average accuracy by 2.5\%.

\subsection{Visualization Comparison with Different Methods}
\label{sec:2_2}

\begin{figure*}[tbp]
  \centering
  \includegraphics[scale=0.32]{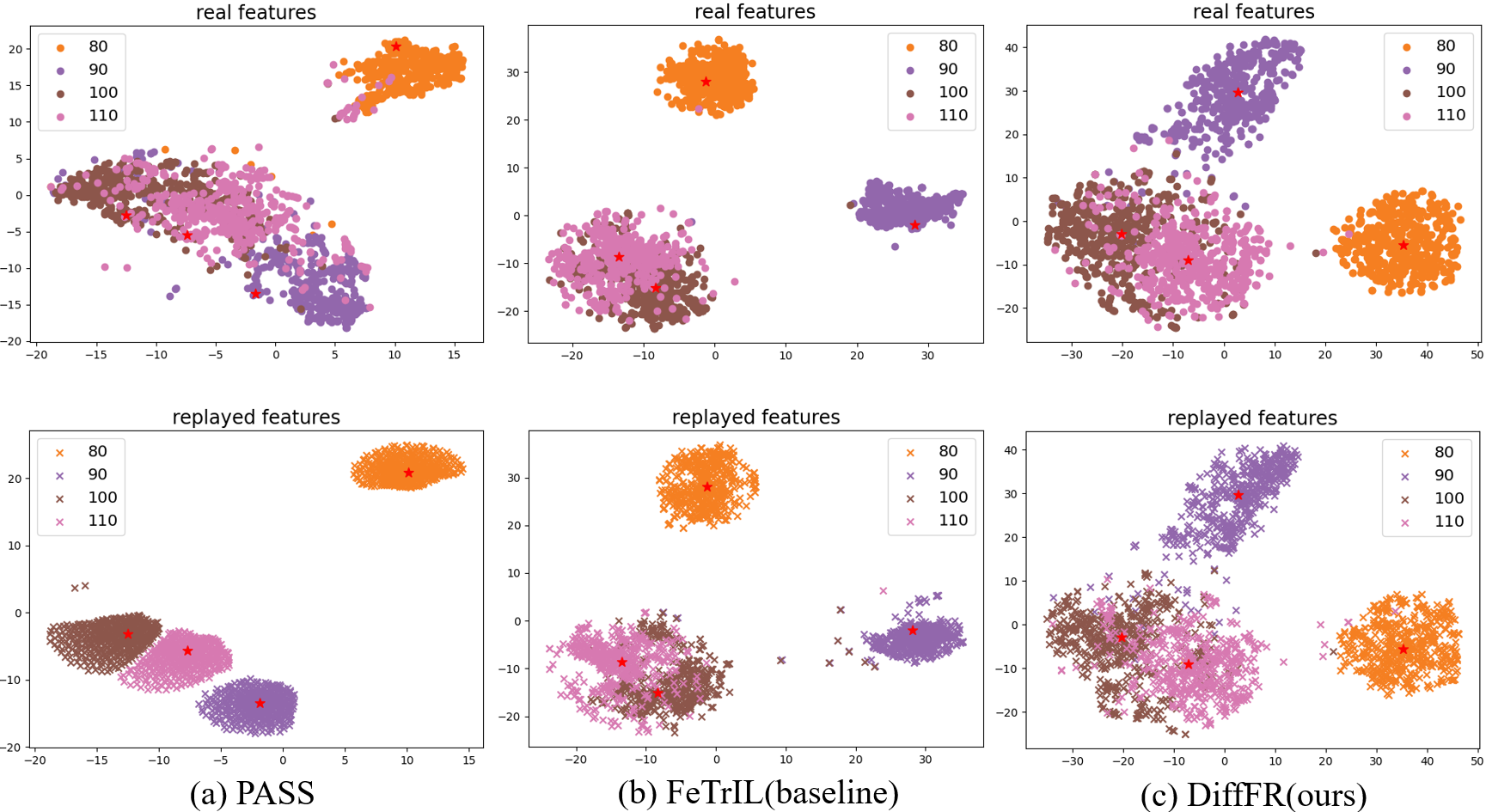}
  \caption{Visualization comparisons of real features and replayed features for PASS, FeTrIL, and DiffFR. The visualization is based on the last stage of TinyImageNet (T=10). Therefore, class 80 and 90 are considered base classes, while class 100 and 110 are considered novel classes. For different methods, we focus on the overlap between real features, as well as the similarity between replayed features and real features. Each class has 500 features and we only visualize 4 classes for clarity.} 
  \label{fig:sup_comp}
  \vspace{-0.2cm}
\end{figure*}



PASS uses distillation to preserve past knowledge and generally favor plasticity, while FeTrIL fixes the feature extractor and generally favor stability. In Fig.\ref{fig:sup_comp}, we visualize the real features and replayed features of PASS, FeTrIL, and DiffFR (ours). On the one hand, the features replayed by PASS and FeTrIL exhibit low similarity from the real features. Generally, there is a more pronounced overlap among features of novel classes (class 100 \& 110), making it crucial to accurately model the distribution shape of novel class features. However, FeTrIL falls short in achieving this, resulting in poor recognition of novel classes. In contrast, diffusion-based feature replay excels in modeling the feature distribution, thereby improving performance.

On the other hand, both PASS and FeTrIL exhibit a larger overlap in novel class features. PASS uses distillation to penalize model updates, restricting its learning capacity for novel classes. FeTrIL only employs supervised learning to train the feature extractor on the initial data and fix it afterwards, leading to limited generalization to unseen new classes. In contrast, DiffFR introduces similarity-based supervision and label augmentation during initial training, promoting the generalization of the feature extractor. Consequently, it reduces the overlap among novel class features, thus improving performance.

\subsection{Comparative Study under Enhanced Data Augmentation}
\label{sec:2_3}
We integrate similarity-based self-supervised learning (SSL) in the initial training, which promotes generalization by bringing closer the features of different views of the same image. Under weak data augmentation used in NECIL, the features of different views are inherently similar, limiting the full demonstration of the advantages of SSL. To further show the effectiveness of SSL in NECIL, we reproduce recent methods (ABD \cite{smith2021always}, PASS \cite{zhu2021prototype}, IL2A \cite{zhu2021class}, SSRE \cite{zhu2022self}, and FeTrIL \cite{petit2023fetril}) under enhanced data augmentation (EA). EA primarily consists of color transformation and cutout, and the reproduction is mainly based on PyCIL \cite{zhou2023pycil}. As shown in Tab.\ref{table:table4}, FeTrIL performs the best among the methods excluding DiffFR. In fact, even considering SOPE \cite{zhu2023self} and PRAKA \cite{shi2023prototype} published at ICCV 2023, FeTrIL is only inferior on CIFAR-100 but still achieves the optimal performance on TinyImageNet and ImageNet-Subset. Therefore, FeTrIL serves as a strong baseline for our method and we focus on the comparison with it.

\begin{table*}[t]
  \centering
  \scalebox{0.78}{
  \begin{tabular}{llllllllll}
    \toprule
    \multirow{2}{*}{\textbf{Methods}}  &
    \multicolumn{3}{c}{CIFAR-100} & \multicolumn{3}{c}{TinyImageNet} & \multicolumn{3}{c}{ImageNet-Subset}  \\
    \cmidrule(r){2-4}
    \cmidrule(r){5-7}
    \cmidrule(r){8-10}
    &  $T$=5     & $T$=10     & $T$=20     & $T$=5     & $T$=10     & $T$=20     & $T$=5     & $T$=10     & $T$=20 \\
    \midrule
    
    ABD \cite{smith2021always}  & 63.8   & 62.5   & 57.4  & -  & -  & -  & -  & -  & -  \\
    PASS \cite{zhu2021prototype}  & 63.5   & 61.8   & 58.1  &  49.6   & 47.3   & 42.1  &  64.4   & 61.8   & 51.3  \\
    IL2A \cite{zhu2021class}  & 66.0   & 60.3   & 57.9  &  47.3   & 44.7   & 40.0  & -  & -  & -  \\
    SSRE \cite{zhu2022self}  & 65.9   & 65.0   & \underline{61.7}  &  50.4   & 48.9   & 48.2  & - & 67.7  & -  \\
    FeTrIL \cite{petit2023fetril}  & \underline{66.3}   & \underline{65.2}   & 61.5  &  \underline{54.8}  & \underline{53.1}    & \underline{52.2}  &  \underline{72.2}   & \underline{71.2}   & \underline{67.1}  \\
    \midrule
    DiffFR (ours) &      
    \textbf{72.2}$^{+5.9}$  & \textbf{71.9}$^{+6.7}$  & \textbf{70.7}$^{+9.0}$  & \textbf{56.4}$^{+1.6}$  & \textbf{55.9}$^{+2.8}$  & \textbf{55.8}$^{+3.6}$  & \textbf{73.7}$^{+1.5}$  & \textbf{73.4}$^{+2.2}$  & \textbf{72.1}$^{+5.0}$  \\
    \midrule
    ABD* \cite{smith2021always}  & 67.0   & 63.2   & 58.7  & -  & -  & -  & -  & -  & -  \\
    PASS* \cite{zhu2021prototype}  & 66.8   & 62.9   & 62.8  &  53.6   & 51.3   & 44.3  &  66.7   & 64.4   & 59.3  \\
    IL2A* \cite{zhu2021class}  & 67.9   & 62.5   & 60.8  &  50.5   & 48.9   & 43.1  & -  & -  & -  \\
    SSRE* \cite{zhu2022self}  & 67.4   & 66.4   & 64.5  &  51.9   & 51.1   & 49.6  & - & 68.1  & -  \\
    FeTrIL* \cite{petit2023fetril}  & \underline{68.1}   & \underline{67.0}   & \underline{64.7}  &  \underline{57.3}  & \underline{56.1}    & \underline{54.8}  &  \underline{73.1}   & \underline{71.9}   & \underline{67.4}  \\
    \midrule
    
    DiffFR* (ours)        &\textbf{72.8}$^{+4.7}$  & \textbf{72.7}$^{+5.7}$  & \textbf{72.1}$^{+7.4}$  & \textbf{62.7}$^{+5.4}$  & \textbf{62.5}$^{+6.4}$  & \textbf{62.5}$^{+7.7}$  & \textbf{76.1}$^{+3.0}$  & \textbf{76.0}$^{+4.1}$  & \textbf{73.9}$^{+6.5}$  \\
   
    \bottomrule
  \end{tabular}}
  \caption{The average incremental accuracy $(\uparrow)$ in NECIL with different incremental phases on CIFAR-100, TinyImageNet, and ImageNet-Subset. We use the superscript to report the improvement over compared results. ”-” means that results are unavailable. \textit{*} means enhanced data augmentation (EA) for initial training. The best and second best results among these methods are in \textbf{bold} and \underline{underlined}.}
  \label{table:table4}
\end{table*}

Without EA, DiffFR outperforms FeTrIL by an average of 7.2\%, 2.7\%, and 2.9\% on CIFAR-100, TinyImageNet, and ImageNet-Subset, respectively. In contrast, the improvements with EA become 5.9\%, 6.5\%, and 4.5\% on the three datasets, respectively. Except for a slight decrease in superiority on CIFAR-100, DiffFR demonstrates a more pronounced advantage on the other two datasets. Taking TinyImageNet as an example, the commonly used data augmentation only includes random horizontal flip and normalization, which is very weak and not conducive to substantiating the superiority of similarity-based SSL. Under weak data augmentation  (without EA), DiffFR exhibits a 2.7\% performance improvement over FeTrIL, and this improvement increases to 6.5\% with EA. While both methods benefit from EA, DiffFR achieves a larger performance improvement, demonstrating the efficacy of similarity-based SSL.

\subsection{Ablation Study on Guidance Scale}
In classifier-free guidance used for class-conditional generation, the guidance scale regulates the trade-off between quality and diversity of the generated data. A larger scale implies higher quality at the expense of diversity. DiffFR excels in effectively modeling the distribution shape, thus minimizing the discrepancy between generated and real features. However, as shown in Tab.\ref{tabel:tabel10}, increasing the guidance scale tends to bias sampling towards the feature center, neglecting crucial boundary information and leading to performance degradation.

\begin{table}[t]
\centering
\scalebox{1}{
\begin{tabular}{p{0.8cm}<{\centering}p{0.8cm}<{\centering}p{0.8cm}<{\centering}p{0.8cm}<{\centering}p{0.8cm}<{\centering}p{0.8cm}<{\centering}p{0.8cm}<{\centering}}
\toprule
scale            & $1$       & $2$      & $3$      & $4$      & $5$     & $6$     \\
\midrule
$A_T$         &\textbf{72.7}         & 72.1     & 71.5        & 71.1         & 70.9         & 70.8         \\
\bottomrule
\end{tabular}}
\caption{The study of guidance scale on CIFAR-100 (10 phases). $A_T$ means the average incremental accuracy.}
\label{tabel:tabel10}

\end{table}

\section{Extensions to Domain-Incremental Learning (DIL)}
\label{sec:3}
Following \cite{shi2024unified}, we perform experiments on two commonly used datasets: Rotating MNIST (R-MNIST) \cite{mnist} and  Sequential CORe50 (Seq-CORe50) \cite{lomonaco2017core50, lomonaco2020rehearsal}. R-MNIST contains 20 sequential domains, with the domain shift being gradual. Specifically, the images in domain $t$ are rotated by an angle randomly sampled from the range $[9^{\circ}\cdot(t-1),9^{\circ}\cdot t)$. CORe50 contains 50 indoor objects collected from 11 domains, totaling 120,000 images. We reserve 20\% of the data for testing and sequentially train the model across these 11 domains. This dataset variant is referred to as Sequential CORe50 (Seq-CORe50). We also report \textit{average incremental accuracy}  and \textit{average forgetting}.

We adopt the same backbone and save no exemplars for replay. Other methods typically reserve a memory size of 500 exemplars in Seq-CORe50, which approximately equates to 6.1M parameters in storage. To better accommodate new knowledge and improve the plasticity, we employ a classifier with a little more parameters (1.2M). The comparison is fair because we do not store any exemplars, resulting in our total memory usage not exceeding that of other methods.

\section{Future Work}
\label{sec:4}
Fixing the feature extractor is simple and effective, and we also fix it. In fact, compared to SOPE and PRAKA with an unfixed feature extractor published in ICCV 2023, FeTrIL with a fixed feature extractor performs better on TinyImageNet and ImageNet-Subset. Similarity-based supervision and diffusion-based feature replay, which are the core components of our method, can be directly applied to distillation-based methods for better plasticity. Therefore, applying them to distillation-based methods is worth exploring, as it may improve the performance of NECIL.

\end{document}